\documentclass[conference]{IEEEtran}
\IEEEoverridecommandlockouts
\usepackage{cite}
\usepackage{amsmath,amssymb,amsfonts}
\usepackage{algorithmic}
\usepackage{graphicx}
\usepackage{textcomp}
\usepackage{xcolor}
\usepackage{orcidlink}
\usepackage{hyperref}
\usepackage{multirow}
\usepackage{graphicx}
\usepackage{amsmath}
\def\BibTeX{{\rm B\kern-.05em{\sc i\kern-.025em b}\kern-.08em
    T\kern-.1667em\lower.7ex\hbox{E}\kern-.125emX}}
\begin{document}

\title{PillarGrid: Deep Learning-based Cooperative Perception for 3D Object Detection from Onboard-Roadside LiDAR
}

\author{
Zhengwei Bai$^{\orcidlink{0000-0002-4867-021X}}$,~\IEEEmembership{Student Member, IEEE},
Guoyuan Wu$^{\orcidlink{0000-0001-6707-6366}}$,~\IEEEmembership{Senior Member, IEEE},
\\Matthew J. Barth$^{\orcidlink{0000-0002-4735-5859}}$,~\IEEEmembership{Fellow, IEEE},
Yongkang Liu,
Emrah Akin Sisbot,
Kentaro Oguchi

\thanks{Zhengwei Bai, Guoyuan Wu, and Matthew J. Barth are with the Department of Electrical and Computer Engineering, the University of California at Riverside, Riverside, CA 92507 USA (e-mail: zbai012@ucr.edu).}

\thanks{Yongkang Liu, Emrah Akin Sisbot, and Kentaro Oguchi are with the Toyota North America R\&D Labs, Mountain View, CA 94043, USA.}

}

\maketitle

\begin{abstract}

3D object detection plays a fundamental role in enabling automated driving, which is regarded as the significant leap forward for contemporary transportation systems from the perspectives of safety, mobility, and sustainability. Most of the state-of-the-art object detection methods from point clouds are developed based on a single onboard LiDAR, whose performance will be inevitably limited by the range and occlusion, especially in dense traffic scenarios. In this paper, we propose \textit{PillarGrid}, a novel cooperative perception method fusing information from multiple 3D LiDARs (both on-board and roadside), to enhance the situation awareness for connected and automated vehicles (CAVs). PillarGrid consists of four main components: 1) cooperative preprocessing of point clouds, 2) pillar-wise voxelization and feature extraction, 3) grid-wise deep fusion of features from multiple sensors, and 4) convolutional neural network (CNN)-based augmented 3D object detection. A novel cooperative perception platform is developed for model training and testing. Extensive experimentation shows that PillarGrid outperforms other single-LiDAR-based 3D object detection methods concerning both accuracy and range by a large margin.

\end{abstract}


\section{Introduction}
    
The rapid progress of intelligent transportation systems (ITS) has greatly promoted the efficiency of daily commuting and goods movement. However, drastically increased number of vehicles has resulted in several major problems in terms of safety, mobility, and sustainability. Taking advantage of advanced perception, wireless communication, and precise control, Cooperative Driving Automation (CDA) has the promise to be the revolutionary solution to the aforementioned challenges~\cite{fagnant2015preparing}.

Deploying CDA in urban environments poses a series of technological challenges. Among others, object detection is arguably the most significant since it lays the perception foundation for other subsequent tasks in CDA~\cite{2021SAE}. To achieve this, high-fidelity sensors such as cameras and LiDARs are implemented to collect data from the surrounding environment. In particular, LiDAR is widely utilized in 3D object detection tasks due to its capability of returning the 3D distance and intensity information with high accuracy, which is crucial in CDA applications~\cite{arnold2019survey}. 

\begin{figure}[!t]
    \centering
    \includegraphics[width=0.5\textwidth]{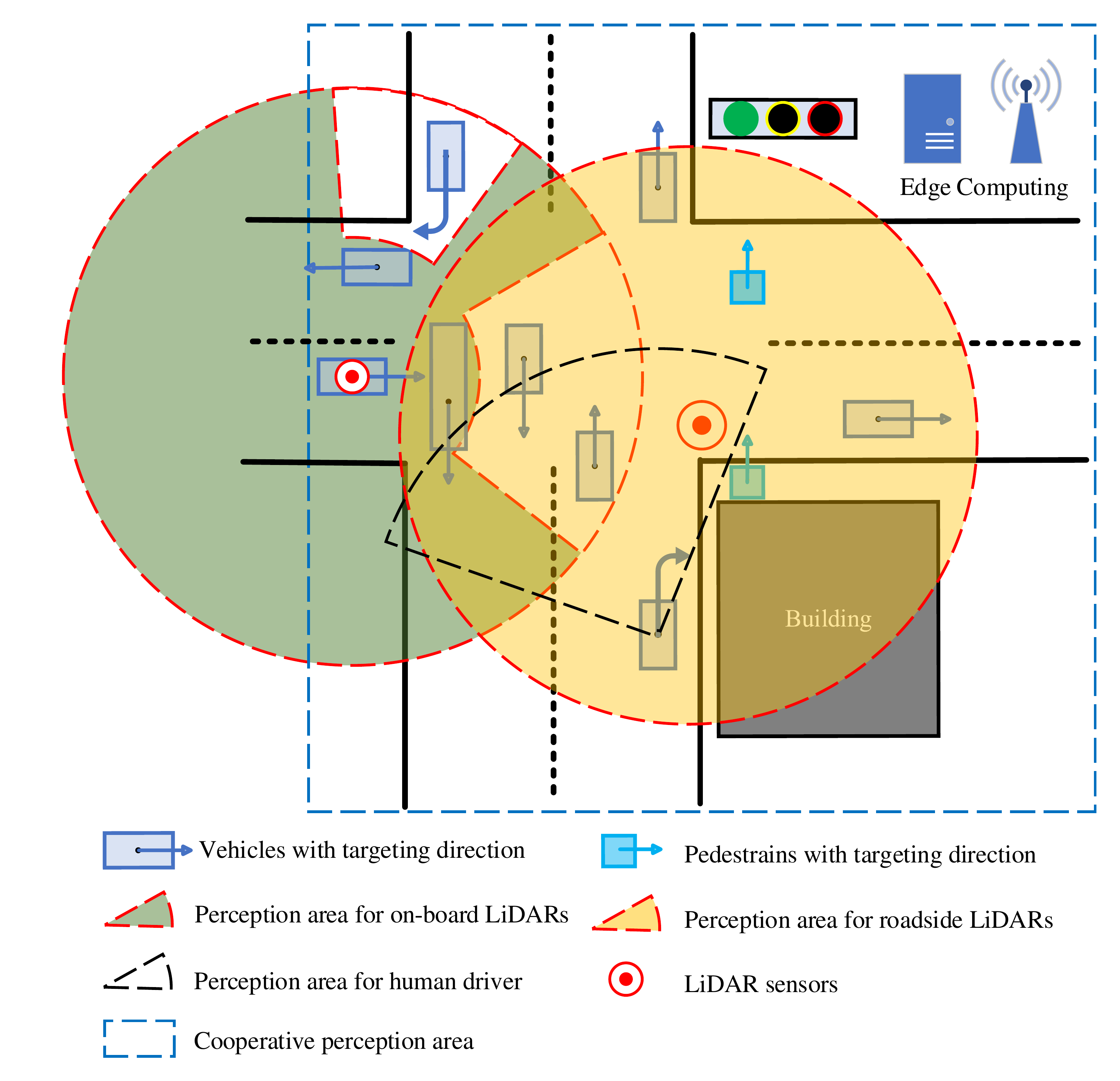}
    \caption{Depiction for the core idea of the proposed cooperative perception method based on the roadside and onboard LiDARs.}
    \label{fig:PillarGridIdea}
\end{figure}

Traditionally, a LiDAR-based object detection pipeline is mainly developed for those 3D LiDARs installed on the top of the equipped vehicle and relies on an onboard computing system to generate the 3D object detection results in real-time~\cite{zamanakos2021comprehensive}.
Following the tremendous success in the application of deep learning methods for computer vision, a large body of literature has explored various algorithms to improve the 3D object detection performance based on point clouds from on-board LiDARs~\cite{lang2019pointpillars, zhou2018voxelnet, Yin_2021_CVPR}. Meanwhile, many onboard sensor-based datasets have been collected and labeled to serve as the foundation for these data-driven methods~\cite{Geiger2012CVPR, caesar2020nuscenes}. However, due to the spatial constraints in the mounting of onboard sensors, their performance is largely limited by the range and occlusion, especially in a dense traffic environment~\cite{bai2022infrastructure}. As a complement, infrastructure-based perception systems have the potential to achieve better object perception results due to fewer occlusion effects and more flexibility in terms of the mounting height and pose.

To further unlock the bottleneck of object detection caused by sensing range and occlusion, Cooperative Perception (CP) is regarded as a promising solution by fusing information from spatially separated sensors~\cite{arnold2020cooperative, marvasti2020cooperative}. In terms of the stage when the information is fused, CP can be divided into three categories: 1) \textit{early-fusion-based} CP (E-CP) which directly runs the perception pipeline on the integrated raw data from multiple sensors; 2) \textit{deep-fusion-based} CP (D-CP) whose fusion process happens within the perception pipeline, i.e., feature-level fusion; and 3) late-fusion-based CP (L-CP) which merges the perception results generated from different pipelines running on respective sensors. E-CP methods need to transmit the raw sensor data which is feasible for sensors with low data volume, e.g., loop detectors and radars. However, it is difficult to apply E-CP methods to LiDAR-based systems due to the limitation of communication bandwidth. Although L-CP methods require much fewer communication capacities, the late-fusion scheme is arguably limited for tapping the potential of cooperative perception to improve detection performance~\cite{arnold2020cooperative}.

In this paper, we propose \textit{PillarGrid}, a novel CP approach that can be applied to both connected and automated vehicles (CAVs) and the roadside infrastructures in terms of LiDAR sensors. Fig.~\ref{fig:PillarGridIdea} depicts the core idea of CP applied in this study, considering both onboard and roadside LiDARs. To be specific, a CAV equipped with an onboard LiDAR can perceive the surrounding environment by itself, covering the area shown in Fig~\ref{fig:PillarGridIdea}. However, due to the occlusion by other vehicles (or buildings), some road users cannot be detected. For connected human-driven vehicles (CHVs), because the pedestrian is blocked by the building, the human driver may not be able to see him/her and take reactions in time, resulting in a risky situation. A roadside LiDAR is usually less impacted by the occlusion, but it still suffers from limited range and/or perception performance degradation due to the decreased resolution with the increase of sensing distance. 

To the best of the authors' knowledge, this paper is the first deep-fusion-based cooperative perception approach that combines point cloud data from an onboard LiDAR and a roadside LiDAR. In this paper, PillarGrid, a novel CP method is proposed to generate 3D object detection based on feature-level cooperation from both onboard and roadside LiDARs. Specifically, Grid-wise Feature Fusion (GFF), a novel deep-fusion method is proposed and a deep neural network is designed to generate the oriented 3D bounding boxes for vehicles and pedestrians. For model training and evaluation, a cooperative perception platform is developed for scenarios design and data acquisition.

The rest of the paper is organized as follows: related work will be reviewed next. Section~\ref{pillar grid} presents the PillarGrid network, followed by the experimental details in Section~\ref{experimental details}. Section~\ref{performance} demonstrates the results and analysis, and the last section concludes this paper with further discussion.

\section{Related Work}
\label{related work}

\subsection{Object Detection in LiDAR Point Clouds}
With the tremendous progress achieved by a convolutional neural network (CNN) in image-based computer vision tasks, CNN is also widely applied to object detection based on point clouds~\cite{zou2019object}. Since LiDAR-based object detection is an intrinsically three-dimensional task, early work~\cite{engelcke2017vote3deep, li20173d} tends to deploy 3D convolutional layers. Although 3D convolutional layers work well for extracting hidden features from 3D point clouds data, they require much higher computational power compared with 2D convolution, resulting in longer processing time. For instance, as a fast object detection model, \textit{Vote3Deep} proposed by Engelcke et al.~\cite{engelcke2017vote3deep} utilized sparse convolutional layers to deal with 3D point clouds and required about $500ms$ to process a single frame of point clouds data, i.e., $2Hz$ inference frequency. Comparatively, the fast 2D object detection models can run at a frequency over $50Hz$~\cite{redmon2016you, liu2016ssd}.

To speed up the 3D object detection, one popular paradigm is to encode the 3D point clouds into a 2D birds-eye-view (BEV) pseudo-image which can be processed by a standard image detection architecture, such as \textit{ComplexYolo}~\cite{simony2018complex} and \textit{PointPillars}~\cite{lang2019pointpillars}. By discretizing point clouds into 3D voxels and then encoding them to a pseudo-image format, these object detectors can run the inference pipeline at a similar frequency to camera-based 2D detectors~\cite{simony2018complex, lang2019pointpillars, redmon2016you, liu2016ssd}.

\subsection{Roadside-LiDAR-based Object Detection}
Although general object detection has gone through a rapid development era, research focusing on roadside sensors is still an emerging topic and has the potential to break the ground for cooperative automated driving, especially in a mixed traffic environment~\cite{gupta2021deep}. Due to the lack of labeled datasets, most of the roadside-LiDAR-based object detectors are still following a traditional paradigm to deal with point clouds, which consists of 1) \textit{Background Filtering} to remove the laser points reflected from the road surface or buildings by applying filtering methods, such as statistics-based background filtering~\cite{wu2017automatic}; 2) \textit{Clustering} to generate clusters for the laser points by implementing clustering methods, such as \textit{Density-Based Spatial Clustering Applications with Noise} (DBSCAN) ~\cite{ester1996density}; and 3) \textit{Classification} to generate different labels for different traffic objects, such as vehicles and pedestrians, based on neural networks~\cite{li2012brief}. This paradigm is straightforward and popular for implementation in the real world~\cite{zhao2019detection}, but has a large performance gap compared with SOTA learning-based detectors in terms of accuracy and real-time performance~\cite{zou2019object}. Although some of the learning-based object detectors trained with on-board datasets can be implemented directly in roadside sensors without further training~\cite{bai2021cmm, bai2022cyber}, the system performance is still inevitably limited by the lack of training data.

\subsection{Cooperative Perception for Object Detection}
Considering the existence of the physical occlusion, it is extremely difficult to further unlock the bottleneck for the detection performance of a single-sensor-based perception system. Perceiving the environment from spatially separated sensors is increasingly becoming a promising solution to address the issues mentioned above~\cite{bai2022infrastructure}.
    
\begin{figure*}[!ht]
    \centering
    \includegraphics[width=\textwidth]{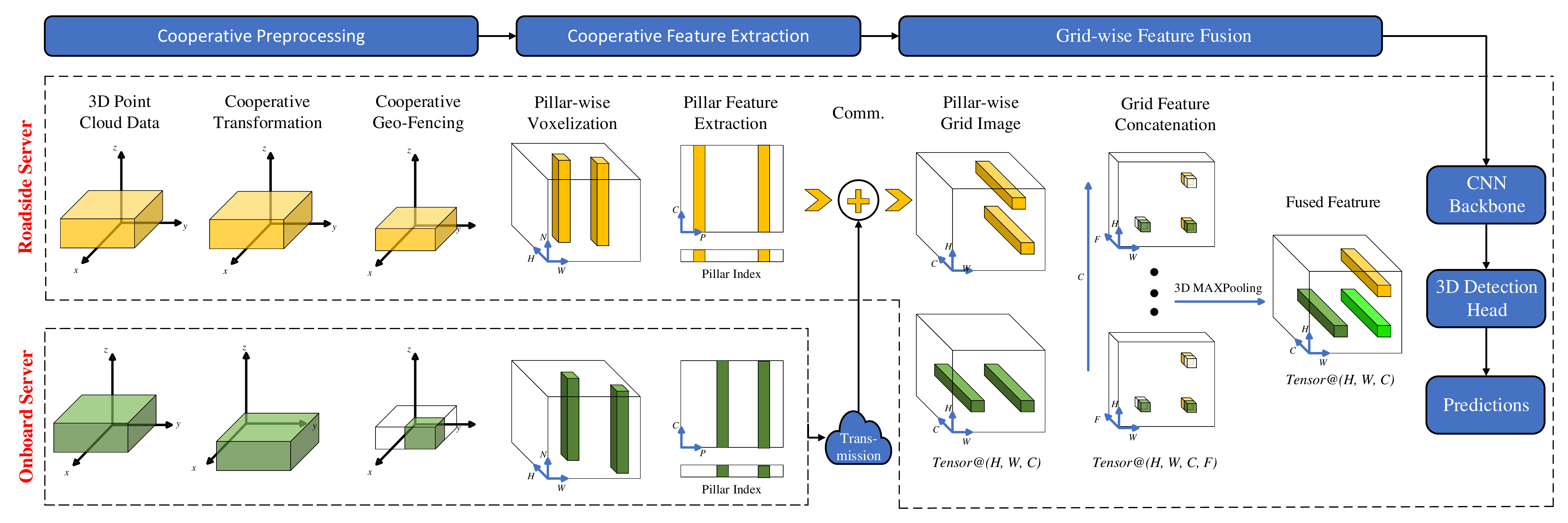}
    \caption{Overview of the PillarGrid Network.}
    \label{fig:pillar grid network}
\end{figure*}
    
Sensor fusion plays the most significant role in a CP system. In terms of the real-world communication bandwidth, the  early-fusion scheme requires high-volume data transmission which is more likely to be satisfied in an infrastructure-based CP system, while late fusion suffers from the detection accuracy~\cite{arnold2020cooperative}. Recently, the deep-fusion scheme is an emerging way to support cooperative object perception by sharing the internal hidden features, which requires less communication power but contains more information than the detection results~\cite{marvasti2020cooperative}.

\section{PillarGrid Network}
\label{pillar grid}
\subsection{Cooperative Preprocessing}
In this paper, the sensors are assumed to be equipped onboard and at the roadside, respectively. Since the two LiDAR sensors are spatially separated, preprocessing has to be done cooperatively in terms of the calibration matrix of these sensors. 
\subsubsection{Cooperative Transformation}
In this paper, the sensor data are collected from a customized CP platform which is introduced later in section~\ref{Dataset Acquisition}. It is to be noted that the calibration matrices for both sensors are designed as the 3D location and rotation information, named \textit{Sensor Location and Pose} (SLaP). Since the point clouds data (PCD) are collected based on a 3D Cartesian coordinate centered with the SLaP of the sensor, cooperative transformation is designed to unify the PCD from different sensors, which is demonstrated in Fig~\ref{fig:pillar grid network}. The raw point cloud data can be described by:
\begin{equation}
    \mathcal{P} = \{[x, y, z, i]^{T} \, |\, [x, y, z]^{T} \in \mathbb{R}^{3}, i\in [0.0, 1.0] \}
\end{equation}
The SLaP information is defined by a six-dimensional vector:
\begin{equation}
    \text{SLaP} = [X, Y, Z, P, \Theta, R]
\end{equation}
where $X$, $Y$, $Z$, $P$, $\Theta$, and $R$ represent the 3D location along $x-$axis, $y-$axis, $z-$axis and the pitch, yaw, and roll angles of the sensors in the global coordinate, respectively.

The process to transform the PCD from sensors' coordinates to the global coordinate is designed by:
\begin{equation}
    \mathcal{R}_{X} = \begin{bmatrix}
1 &  0&  0\\
0 &  cos(-R)&  -sin(-R)\\
0 &  sin(-R)&  cos(-R)
\end{bmatrix}
\end{equation}
\begin{equation}
    \mathcal{R}_{Y} = \begin{bmatrix}
cos(-P) &  0&  sin(-P)\\
0 &  1&  0\\
-sin(-P) &  0&  cos(-P)
\end{bmatrix}
\end{equation}
\begin{equation}
    \mathcal{R}_{Z} = \begin{bmatrix}
cos(-\Theta) &  -sin(-\Theta)&  0\\
sin(-\Theta) &  cos(-\Theta)&  0\\
0 &  0&  1
\end{bmatrix}
\end{equation}
\begin{equation}
\mathcal{T} = [X\text{\space} Y\text{\space}  Z\text{\space} 0]^T    
\end{equation}
\begin{equation}
\mathcal{P}^{S\rightarrow G} = 
\begin{bmatrix}
\mathcal{R}_{X}& 0\\
0&1
\end{bmatrix}
\cdot
\begin{bmatrix}
\mathcal{R}_{Y}& 0\\
0&1
\end{bmatrix}
\cdot
\begin{bmatrix}
\mathcal{R}_{Z}& 0\\
0&1
\end{bmatrix}
\cdot\mathcal{P}^{S} + \mathcal{T}
\end{equation}
where $\mathcal{R}_{X}$, $\mathcal{R}_{Y}$, $\mathcal{R}_{Z}$, and  $\mathcal{T}$ represent the rotation matrix along $x-$axis, $y-$axis, $z-$axis, and the translation matrix, respectively. $\mathcal{P}^{S}$ and $\mathcal{P}^{S\rightarrow G}$ represent the PCD with respect to the sensor's coordinate and the global coordinate, respectively.

\subsubsection{Cooperative Geo-Fencing}
After the cooperative transformation, geo-fencing process is applied to PCD. In this paper, the detected region $\Omega$ for each of the LiDAR sensors is defined as a $102.4m \times 102.4m$ area centered at the location of the respective LiDAR. Specifically, $\mathcal{P}^{S\rightarrow G}$ is geo-fenced by:
\begin{equation}
\mathcal{P}^{S\rightarrow G}_{\Omega} = \{[x, y, z, i]^{T} \, |\, x \in \mathcal{X}, y \in \mathcal{Y}, z \in \mathcal{Z}\}   
\end{equation}
where $P^{S\rightarrow G}_{\Omega}$ represents the 3D point cloud data after geo-fencing; and $\mathcal{X}$ and $\mathcal{Y}$ are set as $[-51.2m, 51.2m]$ for both LiDARs. $\mathcal{Z}$ is set as $[-5.0m, 0m]$ for the roadside LiDAR and $[-3.0m, 2m]$ for the onboard LiDAR.

\subsection{Cooperative Feature Extraction}

\subsubsection{Pillar-wise Voxelization}
To avoid using 3D convolution which is extremely time-consuming, the entire PCD is voxelized  into pillars~\cite{lang2019pointpillars}, shown in Fig.~\ref{fig:pillar grid network}. Both PCD from the roadside LiDAR and the onboard LiDAR are transformed into pillars with the voxel size of $[0.2m, 0.2m, 5.0m]$ in this paper. 
\subsubsection{Pillar Feature Extraction}
To extract the feature simultaneously from two LiDAR sensors, we design a multi-stream neural network as shown in Fig.~\ref{fig:pillar grid network}. For each stream, there are two steps to further solve the PCD.
The first step is geometrical feature extraction, which generates a $D-$dimensional feature vector for every points in pillars, $\mathcal{V}_{p} $ defined by:
\begin{equation}
\mathcal{V}_{p} =\{\left[x, y, z, i, x_{c}, y_{c}, z_{c}, x_{p}, y_{p} \right]_{i}\}^{N}_{i=1}, p = 1, \dots, P
\end{equation}
where $x_{c}, y_{c}, z_{c}, x_{p}, y_{p}, N$, and $P$ represent the distance of each point with respect to the arithmetic center of all points in the $p-$th pillar (the $c$ subscript) and the geometrical center of the pillar (the $p$ subscript), the number of points in the pillar and the number of pillars, respectively. 

Then, to extract the hidden features of $\mathcal{V}_{p}$, a fully connected layer followed by batch normalization is applied. To be specific, the size of $\mathcal{V}_{p}$ is $(D, P, N)$ and the size of the output tensor is $(C, P, N)$, where $C$ is the hidden feature channel and set as $64$ in this paper. Due to the limited communication bandwidth, a data compression method is implemented by selecting the maximum number of features along every channel to synthesize a feature with the size of $(C, P)$, i.e., to compress $N$ points into a $C$-dimensional feature to be transmitted to the edge computer for sensor fusion.

\subsection{Grid-wise Feature Fusion}
Grid-wise Feature Fusion (GFF) is one of the most important stages in PillarGrid, which fuses multi-sensor data based on a mechanism with high expandability. Although only two sensors are deployed in this paper to illustrate our method, GFF can fuse features generated from more than two sensors.

\begin{figure}[!h]
    \centering
    \includegraphics[width=0.5\textwidth]{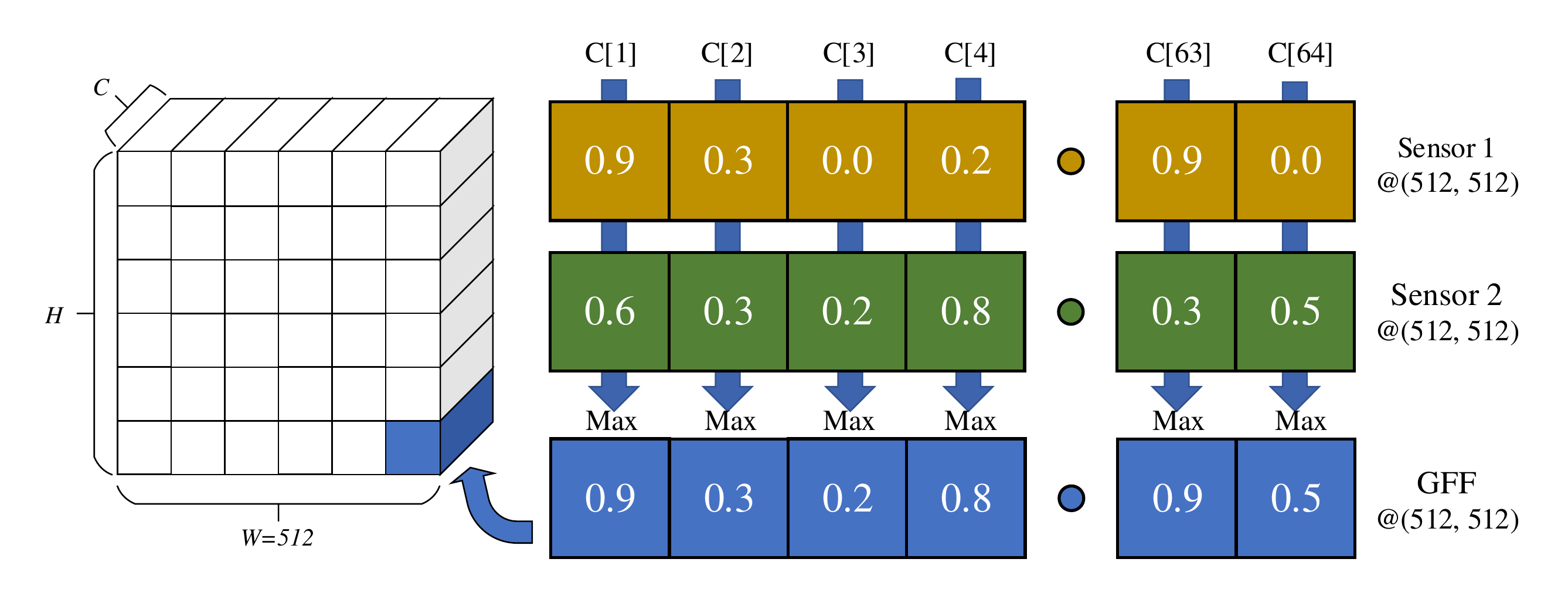}
    \caption{Description of the core process in GFF.}
    \label{fig:GFF}
\end{figure}

Fig.~\ref{fig:GFF} depicts the core process in GFF. The whole process of GFF can be divided into three stages. First, map pillar features with the size of $(C, P)$ into a grid format with the size of $(C, H, W)$ according to the pillar index generated in the voxelization stage. Specifically, $H$ and $W$ are two dimensions of the grid, which are related to the size of voxels and PCD geo-fencing range $\Omega$.

After transforming the feature data into the grid format, each grid contains the hidden features representing the PCD at the specific spatial location. The grid feature from each sensor is then augmented by an additional dimension to enable grid-on-grid fusion by unsqueezing from the size of $(C, H, W)$ to $(C, H, W, 1)$. Then concatenation process is implemented to all the features in terms of the last dimension, i.e., generating the output size of $(C, H, W, F)$ where $F$ is the number of fused sensors, i.e., $F=2$ in this paper.

As demonstrated in Fig.~\ref{fig:GFF}, a \textit{max} operation layer is applied to fuse features over different channels. A 3D max-pooling is implemented to generate the output with the size of $(C, H, W, 1)$, followed by a squeeze operation to transform the size to $(C, H, W)$ which is compatible with 2D CNN backbones.

\subsection{CNN Backbone}
\begin{figure}[!h]
    \centering
    \includegraphics[width=0.5\textwidth]{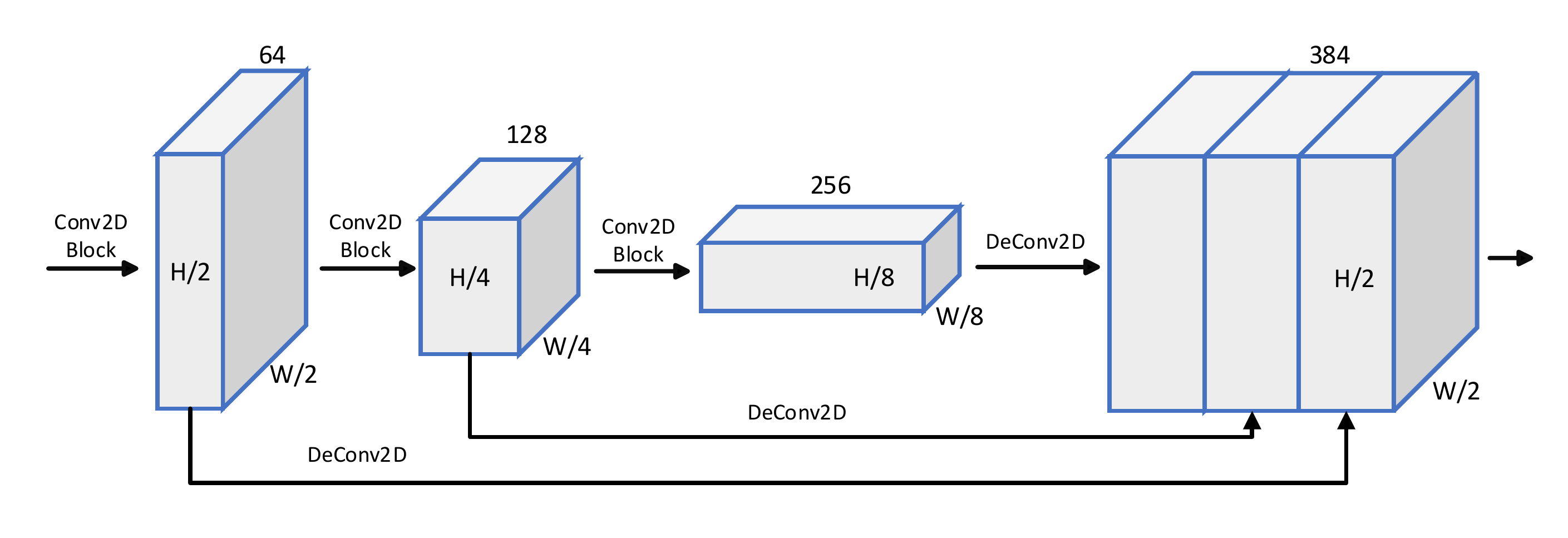}
    \caption{Visualization for the backbone structure.}
    \label{fig:rpn_backbone}
\end{figure}

A Region Proposal Network (RPN)~\cite{zhou2018voxelnet} is applied as the backbone whose structure is shown as Fig.~\ref{fig:rpn_backbone}.
The backbone consists of two sub-networks: 1) one 2D convolutional (Conv2D) layer-based network that generates the extracted features with increasingly small spatial resolution; and 2) one deconvolutional (DeConv2D) layer-based network that generates output features by performing upsampling and concatenation. Each Conv2D block consists of one Conv2D layer with the kernel of $(3, 2, 1)$, followed by several Conv2D layers with kernels of $(3, 1, 1)$. Specifically, the numbers of Conv2D layers in each block are $4$, $6$, and $6$, respectively.

\subsection{3D Detection Head}
An anchor-based 3D detection head~\cite{lang2019pointpillars} is applied to generate 3D bounding boxes. Intersection over Union (IoU) is implemented to match the prior boxes to the ground truth. In this paper, since only vehicles and pedestrians are detected, the number of classes is set to be $2$.

\section{Experimental Details}
\label{experimental details}

\subsection{Loss Function}
Ground truth boxes and anchors are defined by a 7-dimensional vector $(x,y,z,w,l,h,\theta)$ where $w$, $l$, $h$, and $\theta$ represent the width, length, height, and yaw angle, respectively. The loss function is defined as:
\begin{equation}
    \Delta x = \frac{x^{gt} - x^a}{d^a}, 
    \Delta y = \frac{y^{gt} - y^a}{d^a}, 
    \Delta z = \frac{z^{gt} - z^a}{h^a},
\end{equation}
\begin{equation}  
    \Delta w = \log\frac{w^{gt}}{w^a}, 
    \Delta l = \log\frac{l^{gt}}{l^a}, 
    \Delta h = \log\frac{h^{gt}}{h^a}, 
\end{equation}
\begin{equation}  
    \Delta \theta = sin(\theta^{gt} - \theta^a)
\end{equation}
where the superscript $gt$ and $a$ represent the ground truth and anchor, respectively; and $d^a$ is defined by: 
\begin{equation}
    d^a = \sqrt{(w^a)^2 + (l^a)^2}.
\end{equation}
The total localization loss is:
\begin{equation}
    \mathcal{L}_{loc} = \sum_{b \in (x, y, z, w, l, h, \theta)} \text{SmoothL1}(\Delta b)
\end{equation}
The object classification loss is defined as:
\begin{equation}
    \mathcal{L}_{cls} = -\alpha_{a}(1-p^a)^\gamma \log p^a,
\end{equation}
where $p^a$ is the class probability of an anchor; and $\alpha$ and $\gamma$ are set to be $0.25$ and $2$, respectively. Hence, the total loss is:
\begin{equation}
    \mathcal{L} = \frac{1}{N_{pos}}(\beta_{loc}\mathcal{L}_{loc} + \beta_{cls}\mathcal{L}_{cls} + \beta_{dir}\mathcal{L}_{dir}),
\end{equation}
where $N_{pos}$ is the number of positive anchors; and $\beta_{loc}$, $\beta_{cls}$ and $\beta_{dir}$ are set as $2$, $1$, and $0.2$, respectively.

\subsection{Dataset Acquisition}
\label{Dataset Acquisition}
\begin{figure}[!h]
    \centering
    \includegraphics[width=0.4\textwidth]{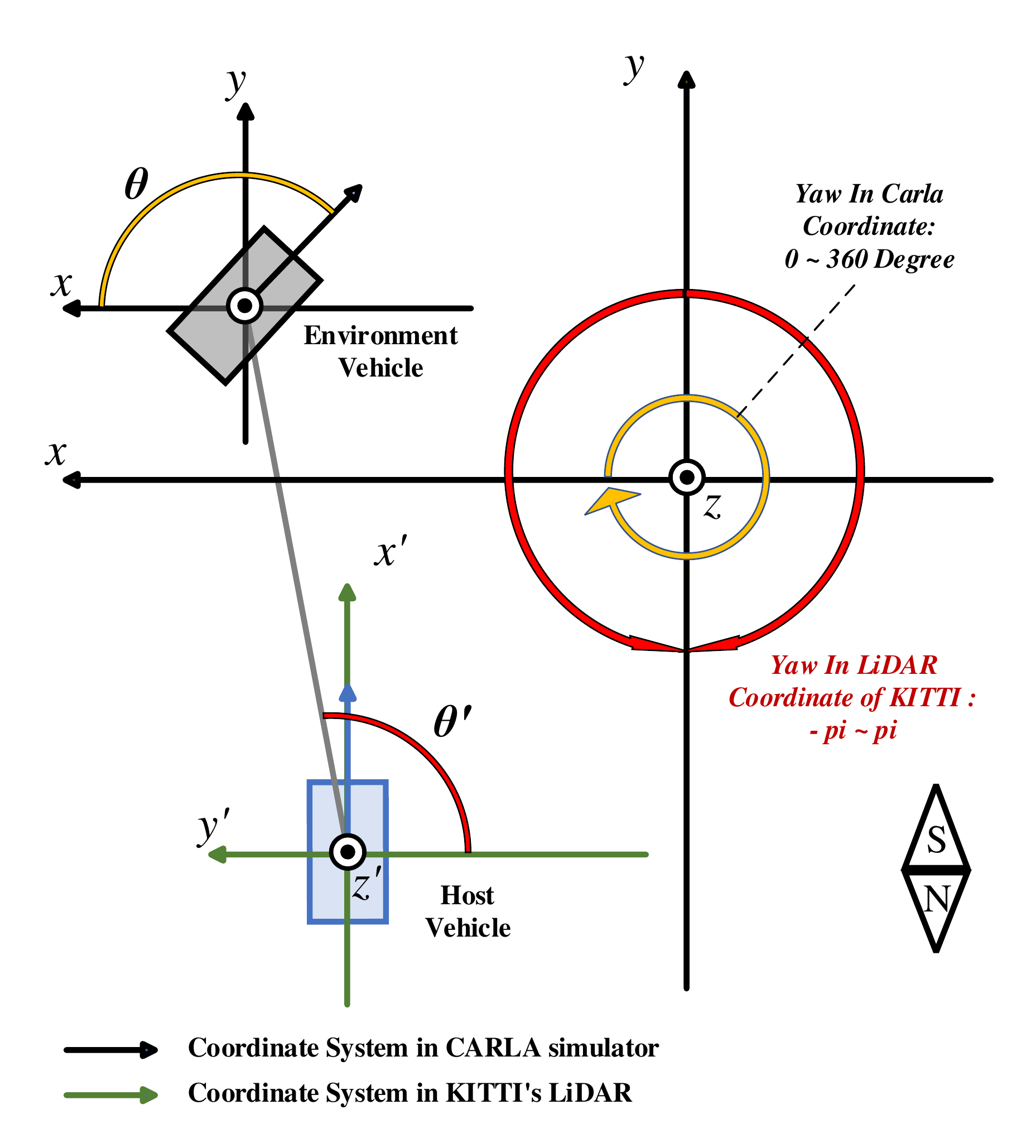}
    \caption{Visualization of data transformation from the CARLA environment to KITTI's format.}
    \label{fig:Carla_kitti_transform}
\end{figure}

Since so far, no dataset is available for supporting cooperative perception (CP) between one onboard LiDAR and one roadside LiDAR, a CP platform is developed to collect sensor data and ground truth labels based on the CARLA simulator~\cite{dosovitskiy2017carla}. In addition, to ease the re-usability and accessibility, we organize the new dataset to be compatible with existing open-source training platforms (e.g., OpenMMLab~\cite{mmdet3d2020}), by following the KITTI's format~\cite{Geiger2012CVPR}. However, the KITTI's settings and CARLA have different coordinate systems, as shown in Fig.~\ref{fig:Carla_kitti_transform}. Therefore, a transformation is designed to generate the ground truth labels in the KITTI coordinate.

\begin{figure}[!t]
    \centering
    \includegraphics[width=0.45\textwidth]{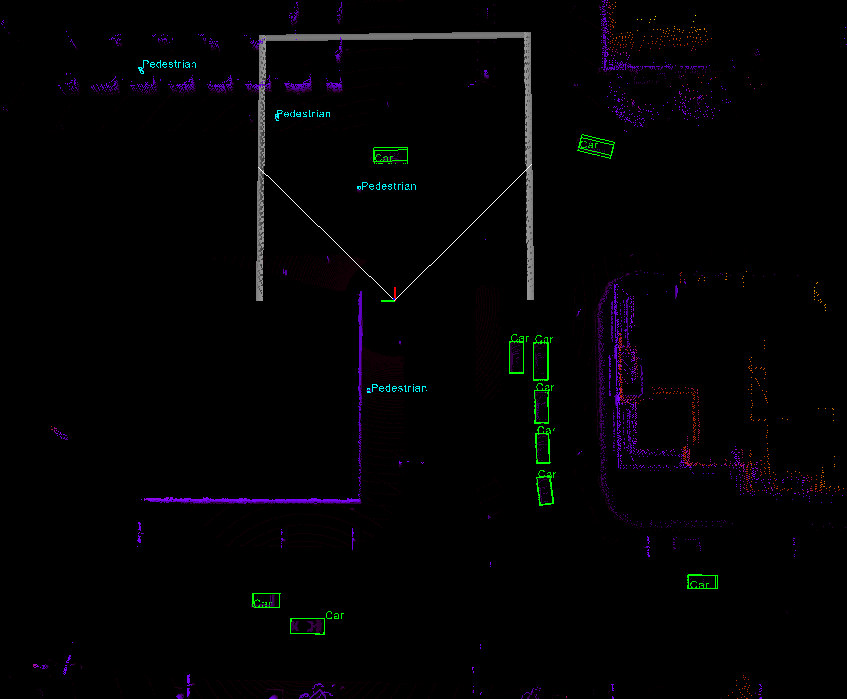}
    \caption{3D point clouds and ground truth labels in one data frame.}
    \label{fig:dataset115515}
\end{figure}

This dataset (named ``\textit{CARTI}'') consists of data from two LiDAR sensors, one is equipped at the roadside (i.e., on a traffic signal pole of an intersection) with the height of $3.74m$, while the other LiDAR is mounted on the top a vehicle with the height of $1.74m$. The simulation is running in a synchronization mode at $10$Hz. Data is collected at $2$Hz, and totally $11,000$ frames of 3D point clouds are available, including $8,000$ for training, $1,000$ for evaluation, and $2,000$ for testing. Fig.~\ref{fig:dataset115515} shows the visualized 3D point clouds and ground truth labels in one data frame.

\section{Results and Analysis}
\label{performance}
The training and testing platform is equipped with Intel$^\text{\textregistered}$ Core™ i7-10700K CPU@3.80GHz×16 and NVIDIA GPU@GeForce RTX 3090. The training pipeline is designed with 160 epochs with \textit{Batchsize} of 10. Two single-sensor-based object detection models based on PointPillar~\cite{lang2019pointpillars}, a SOTA object detection method, are applied to demonstrate the performance of our method.

\subsection{Quantitative Analysis}
Two benchmarks are applied in terms of object detection performance: bird's-eye view (BEV) and 3D perspective. Average precision (AP) and average recall (AR) are measured for each class under the IoU of 0.50 for cars and 0.25 for pedestrians. Mean average precision (mAP) is calculated across all the classes to illustrate the overall performance of the model.

As shown in Table~\ref{tab:eval_BEV} and Table~\ref{tab:eval_3d}, PillarGrid outperforms all baselines in terms of mAP for both BEV and 3D benchmarks. Specifically, for BEV detection, PillarGrid outperforms 44.96\% for onboard detection mAP and 23.72\% for roadside detection mAP. For 3D detection, PillarGrid outperforms 90.52\% for onboard detection mAP and 25.39\% for roadside detection mAP.

\begin{figure*}[!t]
    \centering
    \includegraphics[width=\textwidth]{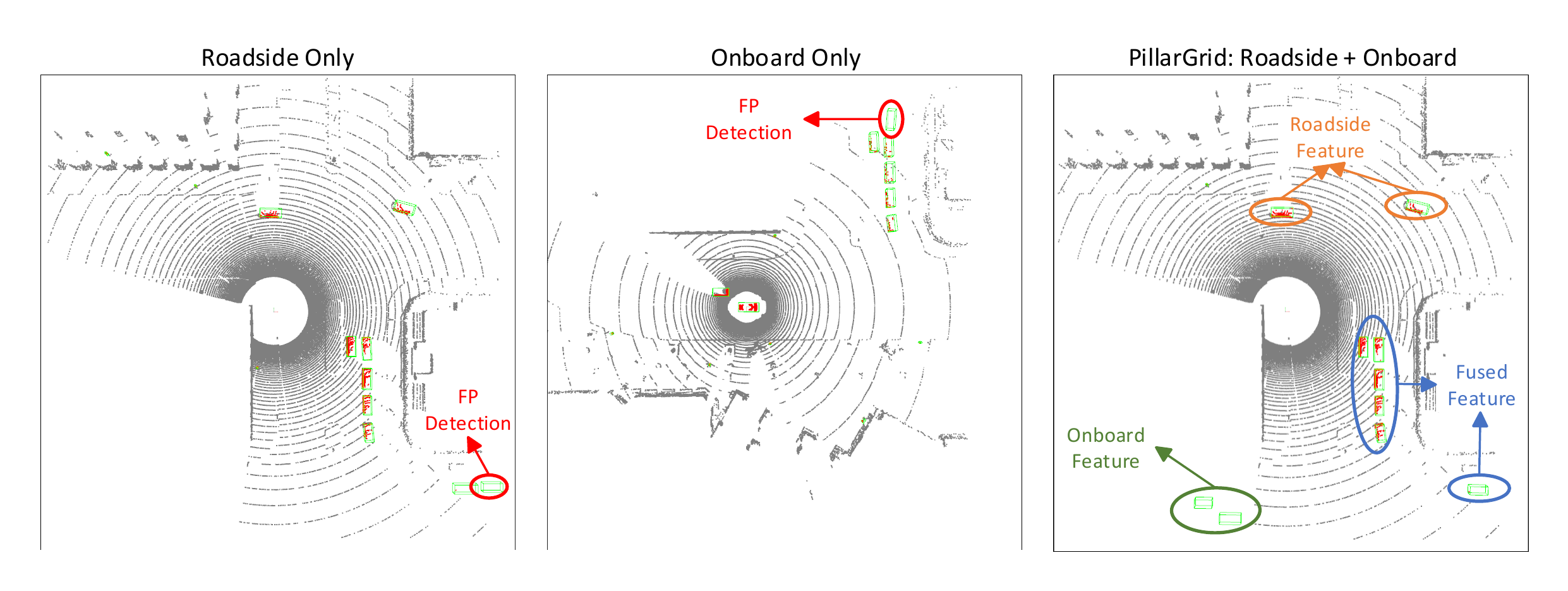}
    \caption{Qualitative analysis on CARTI. We show a BEV of the LiDAR point cloud (gray dots), as well as the 3D bounding boxes (green) for explicit visualization.}
    \label{fig:pillar grid results}
\end{figure*}

For vehicle detection, in both benchmarks, PillarGrid surpasses all the baselines with a large margin in terms of both precision and recall. The results also validate the feasibility of the core idea of the cooperative perception shown in Fig.~\ref{fig:PillarGridIdea}, which represents that PillarGrid can take advantage of both onboard and roadside point clouds. 
For pedestrian detection, PillarGrid outperforms the baselines in terms of the AP but has similar AR performance to roadside detection. According to the Pedestrian AR from onboard detection (lower than 1\%), one hypothesis is that current voxel settings are too large ($0.2m\times 0.2m$) to generate effective pillar features for pedestrians (with the size of $0.38m\times 0.38m$), which may cause the slight degradation of the fused features.

\begin{table}[!h]
\centering
\caption{Results on the CARTI dataset using the BEV detection benchmark.}
\label{tab:eval_BEV}
\resizebox{0.5\textwidth}{!}{%
\begin{tabular}{|c|c||c||cc||cc|}
\hline
\multirow{2}{*}{Method} &
  \multirow{2}{*}{Modality} &
  \multirow{2}{*}{mAP} &
  \multicolumn{2}{c||}{Car} &
  \multicolumn{2}{c|}{Pedestrian} \\ \cline{4-7} 
            &               &       & \multicolumn{1}{c|}{AP50}  & AR50  & \multicolumn{1}{c|}{AP25}  & AR25  \\ \hline
PointPillar &
  Onboard Det. &
  35.96 &
  \multicolumn{1}{c|}{63.63} &
  41.19 &
  \multicolumn{1}{c|}{9.09} &
  0.26 \\
PointPillar & Roadside Det. & 42.03 & \multicolumn{1}{c|}{80.67} & 64.74 & \multicolumn{1}{c|}{28.25} & \textbf{13.16} \\
PillarGrid &
  Cooperative Det. &
  \textbf{52.00} &
  \multicolumn{1}{c|}{\textbf{90.91}} &
  \textbf{85.33} &
  \multicolumn{1}{c|}{\textbf{30.68}} &
  12.70 \\ \hline
\end{tabular}%
}
\end{table}

\begin{table}[!h]
\centering
\caption{Results on the CARTI dataset using the 3D detection benchmark.}
\label{tab:eval_3d}
\resizebox{0.5\textwidth}{!}{%
\begin{tabular}{|c|c||c||cc||cc|}
\hline
\multirow{2}{*}{Method} &
  \multirow{2}{*}{Modality} &
  \multirow{2}{*}{mAP} &
  \multicolumn{2}{c||}{Car} &
  \multicolumn{2}{c|}{Pedestrian} \\ \cline{4-7} 
            &               &       & \multicolumn{1}{c|}{AP50}  & AR50  & \multicolumn{1}{c|}{AP25}  & AR25  \\ \hline
PointPillar &
  Onboard Det. &
  27.11 &
  \multicolumn{1}{c|}{63.61} &
  40.97 &
  \multicolumn{1}{c|}{9.09} &
  0.25 \\
PointPillar & Roadside Det. & 41.19 & \multicolumn{1}{c|}{80.63} & 64.48 & \multicolumn{1}{c|}{26.89} & \textbf{12.42} \\
PillarGrid &
  Cooperative Det. &
  \textbf{51.65} &
  \multicolumn{1}{c|}{\textbf{90.91}} &
  \textbf{85.19} &
  \multicolumn{1}{c|}{\textbf{30.01}} &
  12.23 \\ \hline
\end{tabular}%
}
\end{table}

\subsection{Qualitative Analysis}
To further illustrate the performance of PillarGrid, we provide qualitative analysis results, as shown in Fig.~\ref{fig:pillar grid results}. The left, middle and right subfigures depict the detection results for roadside baseline, onboard baseline, and PillarGrid, respectively, where wrongly detected results are highlighted.

From the perspective of precision, both roadside detection and onboard detection have one wrong detection result marked as False Positive (FP). In the right subfigure of Fig.~\ref{fig:pillar grid results}, i.e., results from PillarGrid, the Grid-wise Feature Fusion process can successfully merge the feature-level information from two spatially separated LiDARs to improve the detection precision (e.g., detection marked as fused feature).

To depict the performance of PillarGrid in terms of recall, we focus on the roadside LiDAR point cloud data in the right subfigure of Fig.~\ref{fig:pillar grid results}. Due to the physical occlusion and sensing range, two vehicles located at the lower-left corner cannot be detected by the roadside LiDAR (due to the building blockage) and another two vehicles on the top of the subfigure cannot be detected by the onboard LiDAR. However, PillarGrid can detect all of these occluded or out-of-range vehicles by fusing the features from both onboard and roadside sensors, which further demonstrates the capability of our deep fusion-based cooperative perception method.



\section{Conclusions and Future Work}
In this paper, we proposed \textit{PillarGrid}, a feature-level cooperative 3D object detection method based on point cloud data (PCD). To the best of our knowledge, this is the first deep learning-based cooperative perception approach that can be applied to PCD from both onboard and roadside LiDARs. The cooperative data preprocess method is designed for point cloud data transformation and geo-fencing. A two-stream neural network is designed to extract and compress hidden features for data transmission. Grid-wise Feature Fusion (GFF), a novel deep-fusion method, is proposed to fuse deep features. To evaluate the proposed method, we develop a CARLA-based cooperative perception platform and generate a cooperative perception dataset named \textit{CARTI}. The results demonstrate that the PillarGrid can improve the performance of 3D object detection by 90.52\% and 25.39\% compared with SOTA detection methods based on point clouds only from an onboard LiDAR or a roadside LiDAR, respectively.

To explore the horizon of PillarGrid in the future, early fusion and late fusion schemes can be integrated to form a hybrid fusion-based perception system to further improve the perception range. Different network configuration can be applied to improve the detection results for pedestrian and other active transportation modes. Furthermore, behavioral analysis can be conducted based on the object-level detection results to support subsequent CDA applications.

\section*{Acknowledgments}
This research was funded by the Toyota Motor North America InfoTech Labs. The contents of this paper reflect the views of the authors, who are responsible for the facts and the accuracy of the data presented herein. The contents do not necessarily reflect the official views of Toyota Motor North America.

\bibliographystyle{IEEEtran}
\bibliography{references}{}

\begin{thebibliography}{10}
\providecommand{\url}[1]{#1}
\csname url@samestyle\endcsname
\providecommand{\newblock}{\relax}
\providecommand{\bibinfo}[2]{#2}
\providecommand{\BIBentrySTDinterwordspacing}{\spaceskip=0pt\relax}
\providecommand{\BIBentryALTinterwordstretchfactor}{4}
\providecommand{\BIBentryALTinterwordspacing}{\spaceskip=\fontdimen2\font plus
\BIBentryALTinterwordstretchfactor\fontdimen3\font minus
  \fontdimen4\font\relax}
\providecommand{\BIBforeignlanguage}[2]{{%
\expandafter\ifx\csname l@#1\endcsname\relax
\typeout{** WARNING: IEEEtran.bst: No hyphenation pattern has been}%
\typeout{** loaded for the language `#1'. Using the pattern for}%
\typeout{** the default language instead.}%
\else
\language=\csname l@#1\endcsname
\fi
#2}}
\providecommand{\BIBdecl}{\relax}
\BIBdecl

\bibitem{fagnant2015preparing}
D.~J. Fagnant and K.~Kockelman, ``Preparing a nation for autonomous vehicles:
  opportunities, barriers and policy recommendations,'' \emph{Transportation
  Research Part A: Policy and Practice}, vol.~77, pp. 167--181, 2015.

\bibitem{2021SAE}
SAE, ``Taxonomy and definitions for terms related to cooperative driving
  automation for on-road motor vehicles j3216\_202005,'' Available:
  \url{https://www.sae.org/standards/content/j3216_202005/}, 2021.

\bibitem{arnold2019survey}
E.~Arnold, O.~Y. Al-Jarrah, M.~Dianati, S.~Fallah, D.~Oxtoby, and
  A.~Mouzakitis, ``A survey on 3d object detection methods for autonomous
  driving applications,'' \emph{IEEE Transactions on Intelligent Transportation
  Systems}, vol.~20, no.~10, pp. 3782--3795, 2019.

\bibitem{zamanakos2021comprehensive}
G.~Zamanakos, L.~Tsochatzidis, A.~Amanatiadis, and I.~Pratikakis, ``A
  comprehensive survey of lidar-based 3d object detection methods with deep
  learning for autonomous driving,'' \emph{Computers \& Graphics}, vol.~99, pp.
  153--181, 2021.

\bibitem{lang2019pointpillars}
A.~H. Lang, S.~Vora, H.~Caesar, L.~Zhou, J.~Yang, and O.~Beijbom,
  ``Pointpillars: Fast encoders for object detection from point clouds,'' in
  \emph{Proceedings of the IEEE/CVF Conference on Computer Vision and Pattern
  Recognition}, 2019, pp. 12\,697--12\,705.

\bibitem{zhou2018voxelnet}
Y.~Zhou and O.~Tuzel, ``Voxelnet: End-to-end learning for point cloud based 3d
  object detection,'' in \emph{Proceedings of the IEEE conference on computer
  vision and pattern recognition}, 2018, pp. 4490--4499.

\bibitem{Yin_2021_CVPR}
T.~Yin, X.~Zhou, and P.~Krahenbuhl, ``Center-based 3d object detection and
  tracking,'' in \emph{Proceedings of the IEEE/CVF Conference on Computer
  Vision and Pattern Recognition (CVPR)}, June 2021, pp. 11\,784--11\,793.

\bibitem{Geiger2012CVPR}
A.~Geiger, P.~Lenz, and R.~Urtasun, ``Are we ready for autonomous driving? the
  kitti vision benchmark suite,'' in \emph{Conference on Computer Vision and
  Pattern Recognition (CVPR)}, 2012.

\bibitem{caesar2020nuscenes}
H.~Caesar, V.~Bankiti, A.~H. Lang, S.~Vora, V.~E. Liong, Q.~Xu, A.~Krishnan,
  Y.~Pan, G.~Baldan, and O.~Beijbom, ``nuscenes: A multimodal dataset for
  autonomous driving,'' in \emph{Proceedings of the IEEE/CVF conference on
  computer vision and pattern recognition}, 2020, pp. 11\,621--11\,631.

\bibitem{bai2022infrastructure}
Z.~Bai, G.~Wu, X.~Qi, Y.~Liu, K.~Oguchi, and M.~J. Barth,
  ``Infrastructure-based object detection and tracking for cooperative driving
  automation: A survey,'' \emph{arXiv preprint arXiv:2201.11871}, 2022.

\bibitem{arnold2020cooperative}
E.~Arnold, M.~Dianati, R.~de~Temple, and S.~Fallah, ``Cooperative perception
  for 3d object detection in driving scenarios using infrastructure sensors,''
  \emph{IEEE Transactions on Intelligent Transportation Systems}, 2020.

\bibitem{marvasti2020cooperative}
E.~E. Marvasti, A.~Raftari, A.~E. Marvasti, Y.~P. Fallah, R.~Guo, and H.~Lu,
  ``Cooperative lidar object detection via feature sharing in deep networks,''
  in \emph{2020 IEEE 92nd Vehicular Technology Conference
  (VTC2020-Fall)}.\hskip 1em plus 0.5em minus 0.4em\relax IEEE, 2020, pp. 1--7.

\bibitem{zou2019object}
Z.~Zou, Z.~Shi, Y.~Guo, and J.~Ye, ``Object detection in 20 years: A survey,''
  \emph{arXiv preprint arXiv:1905.05055}, 2019.

\bibitem{engelcke2017vote3deep}
M.~Engelcke, D.~Rao, D.~Z. Wang, C.~H. Tong, and I.~Posner, ``Vote3deep: Fast
  object detection in 3d point clouds using efficient convolutional neural
  networks,'' in \emph{2017 IEEE International Conference on Robotics and
  Automation (ICRA)}.\hskip 1em plus 0.5em minus 0.4em\relax IEEE, 2017, pp.
  1355--1361.

\bibitem{li20173d}
B.~Li, ``3d fully convolutional network for vehicle detection in point cloud,''
  in \emph{2017 IEEE/RSJ International Conference on Intelligent Robots and
  Systems (IROS)}.\hskip 1em plus 0.5em minus 0.4em\relax IEEE, 2017, pp.
  1513--1518.

\bibitem{redmon2016you}
J.~Redmon, S.~Divvala, R.~Girshick, and A.~Farhadi, ``You only look once:
  Unified, real-time object detection,'' in \emph{Proceedings of the IEEE
  conference on computer vision and pattern recognition}, 2016, pp. 779--788.

\bibitem{liu2016ssd}
W.~Liu, D.~Anguelov, D.~Erhan, C.~Szegedy, S.~Reed, C.-Y. Fu, and A.~C. Berg,
  ``Ssd: Single shot multibox detector,'' in \emph{European conference on
  computer vision}.\hskip 1em plus 0.5em minus 0.4em\relax Springer, 2016, pp.
  21--37.

\bibitem{simony2018complex}
M.~Simony, S.~Milzy, K.~Amendey, and H.-M. Gross, ``Complex-yolo: An
  euler-region-proposal for real-time 3d object detection on point clouds,'' in
  \emph{Proceedings of the European Conference on Computer Vision (ECCV)
  Workshops}, 2018, pp. 0--0.

\bibitem{gupta2021deep}
A.~Gupta, A.~Anpalagan, L.~Guan, and A.~S. Khwaja, ``Deep learning for object
  detection and scene perception in self-driving cars: Survey, challenges, and
  open issues,'' \emph{Array}, p. 100057, 2021.

\bibitem{wu2017automatic}
J.~Wu, H.~Xu, and J.~Zheng, ``Automatic background filtering and lane
  identification with roadside lidar data,'' in \emph{2017 IEEE 20th
  International Conference on Intelligent Transportation Systems (ITSC)}.\hskip
  1em plus 0.5em minus 0.4em\relax IEEE, 2017, pp. 1--6.

\bibitem{ester1996density}
M.~Ester, H.-P. Kriegel, J.~Sander, X.~Xu \emph{et~al.}, ``A density-based
  algorithm for discovering clusters in large spatial databases with noise.''
  in \emph{kdd}, vol.~96, no.~34, 1996, pp. 226--231.

\bibitem{li2012brief}
J.~Li, J.-h. Cheng, J.-y. Shi, and F.~Huang, ``Brief introduction of back
  propagation (bp) neural network algorithm and its improvement,'' in
  \emph{Advances in computer science and information engineering}.\hskip 1em
  plus 0.5em minus 0.4em\relax Springer, 2012, pp. 553--558.

\bibitem{zhao2019detection}
J.~Zhao, H.~Xu, H.~Liu, J.~Wu, Y.~Zheng, and D.~Wu, ``Detection and tracking of
  pedestrians and vehicles using roadside lidar sensors,'' \emph{Transportation
  research part C: emerging technologies}, vol. 100, pp. 68--87, 2019.

\bibitem{bai2021cmm}
Z.~Bai, S.~P. Nayak, X.~Zhao, G.~Wu, M.~J. Barth, X.~Qi, Y.~Liu, and K.~Oguchi,
  ``Cyber mobility mirror: Deep learning-based real-time 3d object perception
  and reconstruction using roadside lidar,'' \emph{arXiv preprint
  arXiv:2202.13505}, 2022.

\bibitem{bai2022cyber}
Z.~Bai, G.~Wu, X.~Qi, K.~Oguchi, and M.~J. Barth, ``Cyber mobility mirror for
  enabling cooperative driving automation: A co-simulation platform,''
  \emph{arXiv preprint arXiv:2201.09463}, 2022.

\bibitem{dosovitskiy2017carla}
A.~Dosovitskiy, G.~Ros, F.~Codevilla, A.~Lopez, and V.~Koltun, ``Carla: An open
  urban driving simulator,'' in \emph{Conference on robot learning}.\hskip 1em
  plus 0.5em minus 0.4em\relax PMLR, 2017, pp. 1--16.

\bibitem{mmdet3d2020}
M.~Contributors, ``{MMDetection3D: OpenMMLab} next-generation platform for
  general {3D} object detection,''
  \url{https://github.com/open-mmlab/mmdetection3d}, 2020.

\end{thebibliography}

\end{document}